# Comparative Study of Statistical Skin Detection Algorithms for Sub-Continental Human Images

M. R. Tabassum, A. U. Gias, M. M. Kamal, H. M. Muctadir, M. Ibrahim, A. K. Shakir, A. Imran, S. Islam, M. G. Rabbani[1], S. M. Khaled, M. S. Islam, Z. Begum

Institute of Information Technology, University of Dhaka, Dhaka-1000, Bangladesh
[1]Department of Statistics, Biostatistics and Informatics, University of Dhaka, Dhaka-1000, Bangladesh

**Abstract:** *Object detection has been a focus of research in human-computer interaction. Skin area detection has been a key to different recognitions like face recognition, human motion detection, pornographic and nude image prediction, etc. Most of the research done in the fields of skin detection has been trained and tested on human images of African, Mongolian and Anglo-Saxon ethnic origins. Although there are several intensity invariant approaches to skin detection, the skin color of Indian sub-continentals have not been focused separately. The approach of this research is to make a comparative study between three image segmentation approaches using Indian sub-continental human images, to optimize the detection criteria, and to find some efficient parameters to detect the skin area from these images. The experiments observed that HSV color model based approach to Indian sub-continental skin detection is more suitable with considerable success rate of 91.1% true positives and 88.1% true negatives.*

**Key words:** image processing, color space model, color segmentation, skin detection.

## 1. Introduction

In this era of the Internet and multimedia computing, there is a high demand for content based retrieval – the methodology of determining some object in a large collection that depict some particular types of properties. There is a keen need for systems that can estimate the content of an image automatically based on image information. Skin detection plays an important role in tracking people, filtering out adult web images, and facilitating Human Computer Interaction.

The main challenge in skin detection is to make the recognition robust to the large variations in appearance of skin that may occur, like in color and shape, effects of occultation, intensity, color, location of light source, etc. Imaging noise can appear as speckles of skin like color, and many other objects like wood, cooper and some clothes are often confused as skin.

In general human skin is characterized by a combination of red and melanin (yellow, brown) and there is somewhat a range of hue for skin and saturation that represent skin-like pixels. More deeply colored skins are with more melanin, the saturation is more when the skin is yellowish (Rossotti, 1983). The main goal of skin detection and classification is to build a decision rule that discriminate between skin and non-skin pixels. Identifying skin color pixels involves finding a range of values for which most skin pixels would fall in a given color space. The target is to achieve a high detection rate and low false positive rate, that is, skin pixels must be detected in maximum and the amount of non-skin pixels classified as skin should be minimized.

Most works done in the area of skin detection have been concentrated on detecting skins of European, Black or East Asian ethnicities, whereas less focus have been concentrated to detect Indian-like skins. This paper makes a comparative study of three algorithms that use three different color models for statistical skin detection. For training and testing of algorithms used in this study, 200 skin and non skin images of people from Indian sub-continent some of which have been captured locally and others collected randomly from the Internet have been used. The prime focus of experimentation is to find out image properties that are best suited for searching Indian skin in images with cluttered background.

The paper is organized as follows: section 2 covers the relevant works on skin detection, section 3 describes the algorithms used, section 4 describes the experimental setup and makes an analysis of performance of given algorithms and section 5 makes the concluding remarks.

## 2. Related Works

Human skin color has been used to identify and differentiate the skin. This has been proven as useful methods applicable in face recognition, identification of nude and pornographic images (Fleck *et al.*, 1996) (Jones and Rehg, 2002) and also such image processing tasks have been used extensively by intelligence agencies (Ahlberg, 1999).

A number of image processing models have been applied for skin detection. The major paradigms included heuristic and recognizing patterns which were used to obtain accurate results. Among various types of skin detection methods, the ones that make use of the skin color as a tool for the detection of skin is considered to be the most effective (Zarit *et al.*, 1999). Human skins have a characteristic color and it was a commonly accepted idea driven by logic to design a method based on skin color identification. The problem arose with the provision of different varieties of human skin found in different parts of the world. A number of published researches included various skin models and detection techniques (Zarit *et al.*, 1999) (Terrillon *et al.*, 2000) (Brand and Mason, 2000), however, none came up with complete accuracy.

There have been many problematic issues in the domain of skin detection. The choice of color space, the model of precise skin color distribution, and the way of mechanizing color segmentation research for the detection of human skin. Most researches have been focused on pixel based skin recognition, classifying each pixel either as skin or non skin. Each pixel is considered to be an individual unit (Brand and Mason, 2000). Pixel-base skin recognition is considered to be one of the finest models that under normal conditions gives high level of accuracy at the detection phase of the process. Due to its high applicability and efficiency, some color models are used extensively in the arena of skin detection. These models make use of pixel based skin recognition using a model such as RGB (Albiol et. al., 2001). The RGB model makes use of the three colors red, green and blue to identify the chrominance. Then using an efficient model the necessary range is used and applied to a selected photo to determine the presence of skin. Skin color varies depending on ethnicity and region, therefore additional work must to address the issue.

There is another model called the region based method. This method was applied by (Kruppa, 2002) (Yang and Ahuja, 1999) (Gomez and Morales, 2002). In this method the researchers considered the spatial method of skin pixels, and took them into account during the detection phase with the target of maximizing efficiency. As a contradiction to the fact that different people have different types of skin colors, it is found that the major difference does not lie in their chrominance; rather it is determined by intensity to a large extent. The simplest color models useful for intensity invariant skin detection are HSV (Lee and Yoo, 2002), YUV (Chai and Bouzerdoum, 2000) or YIQ (Martinkauppi and Soriano, 2001). The HSV model is an effective mechanism to determine human skin based on hue and saturation. Other efficient models are YUV and YIQ which follow the same brand of modeling using the RGB color space.

This paper mainly focuses on the use of YUV, YIQ and HSV models to determine skin. It also focuses on the effectiveness of the models discussed above. It focuses on the comparative study of the models to determine its effectiveness with respect to Indian sub-continental skin.

Fig. 2.1(a) depicts the color cluster of hue for Indian sub-continental skin. This has been found by plotting the hue values of a number of randomly chosen human skin images. Fig. 2.1(b) and 2.1(c) shows the Indian sub-continental skin color clusters using YUV and YIQ color models.

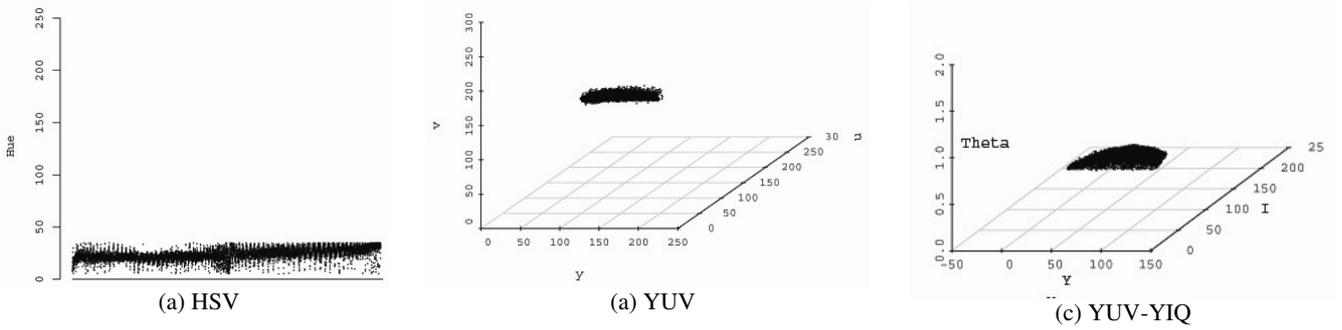

(a) HSV        (a) YUV        (c) YUV-YIQ

**Fig. 2.1:** Indian sub-continental Skin Color Cluster in Different Color Models

## 3. Color Models for Skin Detection

The skin color detection significantly depends on the chosen color model. The RGB color space is default in many image formats. Color space transformation can be applied to reduce the overlap between skin and non-skin pixels and will thereby aid skin pixel classification and achieve high accuracy in varying illumination conditions.

### 3.1 The HSV Color Space

Hue-saturation based color spaces were introduced when there was a need for the user to specify color properties numerically. *Hue* defines the dominant color (such as red, green, purple and yellow) of an area; *saturation* measures the colorfulness of an area in proportion to its brightness. The "intensity", "lightness" or "value" is related to the color luminance (Vezhnevets *et. al.*, 2003). Hue can be used as a decision parameter to detect human skin. Algorithm 3.1 presents the HSV color model based skin detection procedure.

### 3.2 The YUV Color Space

YUV is the color space used in the PAL system of television broadcasting which is the standard in most of Europe and some other places. The RGB values are transformed into YUV values using the formulation given below:

$$\begin{bmatrix} Y \\ U \\ V \end{bmatrix} = \begin{bmatrix} 16 \\ 128 \\ 128 \end{bmatrix} + \begin{bmatrix} 0.257 & 0.504 & 0.098 \\ -0.148 & -0.291 & 0.439 \\ 0.439 & -0.368 & -0.071 \end{bmatrix} \begin{bmatrix} R \\ G \\ B \end{bmatrix}$$

The chromaticity information is encoded in the *U* and *V* components (Bourke, 1994). Hue and saturation are gotten by the following transformation.

$$ch = \sqrt{|U|^2 + |V|^2} \quad \text{and} \quad \theta = \tan^{-1}(|V|/|U|)$$

θ represents hue, which is defined as the angle of vector in *YUV* color space. *Ch* represents saturation, which is defined as the mode of *U* and *V* (Bourke, 1994). Algorithm 3.2 presents the YUV color model based skin detection procedure.

```
Algorithm 3.1: SkinHSV( )
Input Parameters
        Image: Input Image
        T₁ and T₂: Thresholds, discussed in Section 4
Input Parameters
        Image: Image with skin pixels detected
Procedure
    1.  Read the image header, BMPhead
    2.  for i = 1 to BMPhead.height
    3.      for j = 1 to BMPhead.width
    4.          Read pixel Color
    5.          mx = max(Color·R, Color·G, Color·B)
    6.          mn = min(Color·R, Color·G, Color·B)
    7.          ∂ = mx − mn
    8.          If (mx = Color·R) then
    9.              h = (Color·G − Color·B)/δ
    10.         Else If (mx = Color·G) then
    11.             h = 2 + (Color·B − Color·R)/δ
    12.         Else
    13.             h = 4 + (Color·R − Color·G)/δ
    14.         End if
    15.         h = h × 60
    16.         If(h<0)  then
    17.             h = h + 360
    18.         End if
    19.         if( T₁ ≤ h ≤ T₂ ) then
    20.             detect pixel as skin
    21.         end if
    22.     end
    23. end
```

```
Algorithm 3.2: SkinYUV( )
Input Parameters
        Image: Input Image
        T₁ and T₂: Thresholds, discussed in Section 4
Input Parameters
        Image: Image with skin pixels detected
Procedure
    1.  Read the image header, BMPhead
    2.  for i = 1 to BMPhead.height
    3.      for j = 1 to BMPhead.width
    4.          Read pixel Color
    5.          y = 0.257×Color·R + 0.504×Color·G + 0.098×Color·B + 16
    6.          u = −0.148×Color·R − 0.291×Color·G + 0.439×Color·B + 128
    7.          v = 0.439×Color·R − 0.368×Color·G − 0.071×Color·B + 128
    8.          If ( T₁ ≤ y ≤ T₂ & T₃ ≤ u ≤ T₄ & T₅ ≤ v ≤ T₆ ) then
    9.              detect pixel as skin
    10.             end if
    11.     end
    12. end
```

### 3.3 The YIQ Color Space

Like *YUV* color space, *YIQ* is the color primary system adopted by NTSC for color TV broadcasting. Conversion from *RGB* to *YIQ* may be accomplished using the color matrix:

$$\begin{bmatrix} Y \\ I \\ Q \end{bmatrix} = \begin{bmatrix} 0.299 & 0.587 & 0.114 \\ 0.596 & -0.274 & -0.322 \\ 0.211 & -0.523 & 0.312 \end{bmatrix} \begin{bmatrix} R \\ G \\ B \end{bmatrix}$$

*I* is the red-orange axis, *Q* is roughly orthogonal to *I*. The less *I* value means the less blue-green and the more yellow (Bourke, 1994). Through some experiments, we find that the combination of *YUV* and *YIQ* color space is more robust than each other. Algorithm 3.3 presents the YUV and YIQ color model based skin detection procedure.

```
Algorithm 3.3: SkinYUV_YIQ( )
Input Parameters
    Image: Input Image
    T₁, T₂, T₃, T₄, T₅ and T₆: Thresholds, discussed in Section 4
Input Parameters
    Image: Image with skin pixels detected
Procedure
  1. Read the image header, BMPhead
  2. for i = 1 to BMPhead.height
  3.   for j = 1 to BMPhead.width
  4.     Read pixel Color
  5.     y = 0.257×Color·R + 0.504×Color·G + 0.098×Color·B + 16
  6.     u = −0.148×Color·R − 0.291×Color·G + 0.439×Color·B + 128
  7.     v = 0.439×Color·R − 0.368×Color·G − 0.071×Color·B + 128
  8.     θ = tan⁻¹(v/u)
  9.     I = 0.596×Color·R − 0.274×Color·G − 0.322×Color·B
 10.     If ( T₁ ≤ y ≤ T₂ & T₃ ≤ I ≤ T₄ & T₅ ≤ θ ≤ T₆ ) then
 11.       detect pixel as skin
 12.     end if
 13.   end
 14. end
```

## 4. Experimental Setup and Performance Analysis

The comparative study presented in this paper measures the human images of Indian sub-continental region. It considered 200 color images for training and testing of given algorithms and evaluate performance. Among these images there were 60 images where all the image area was covered by skin, 70 images did not have any skin and the other 70 had both human skin and other objects. A total of 120 images were used in the training phase of the experimentation, while the other 80 was used to test the performance. The data acquisition has been done in two ways. A digital camera was used to capture some human and non-human images, while other images for testing and training has been randomly collected from the Internet. The images used for training and testing were under different lighting and illumination conditions.

The first challenge in the study is to find out effective range for thresholds of decision parameters (as discussed in previous section) on which the detection of skin depends. Different statistical tools of central tendency and standard deviation have been used to reach some initial ranges of threshold used for skin recognition. Algorithms with these initial thresholds were trained on a collection of test images. The range of these thresholds was heuristically modified to optimize the value range and get better detection rate. Table 4.1, 4.2, 4.3 provides instances such threshold optimization with respect to False Positives, True positives, True negatives and False negatives.

**Table 4.1:** Threshold Optimization for Detection based on HSV Color Model

| Range | True Positive | True Negative | False Positive | False Negative |
|---|---|---|---|---|
| 2<h<45 | 95.4 | 81.8 | 18.2 | 4.6 |
| 4<h<40 | 93.2 | 83.6 | 16.4 | 6.8 |
| 5<h<35 | 91.1 | 88.1 | 11.9 | 8.9 |
| 10<h<30 | 84.8 | 90.2 | 9.8 | 15.2 |

**Table 4.2:** Threshold Optimization for Detection based on YUV Color Model

| Range | True Positive | True Negative | False Positive | False Negative |
|---|---|---|---|---|
| 75<y<185 & 105<u<150 & 100<v<180 | 92.2 | 81.2 | 18.8 | 12.8 |
| 70<y<175 & 95<u<145 & 95<v<170 | 91.4 | 84.4 | 15.6 | 8.6 |
| 65<y<170 & 85<u<140 & 85<v<160 | 88.3 | 88.4 | 11.6 | 11.7 |
| 60<y<160 & 80<u<135 & 75<v<150 | 86.8 | 91.3 | 8.7 | 13.2 |

Table 4.3: Threshold Optimization for Detection based on YUV-YIQ Color Model

| Range | True Positive | True Negative | False Positive | False Negative |
|---|---|---|---|---|
| 75<y<190 &10<I<122 & -60< θ <170 | 96.8 | 75.5 | 24.5 | 3.2 |
| 75<y<185 &15<I<112 & -54< θ <160 | 94.5 | 77.3 | 22.7 | 5.5 |
| 70<y<175 & 20 <I<102 & -48< θ <150 | 91.2 | 81.5 | 18.5 | 8.8 |
| 65<y<170 & 25 <I<102 & -42< θ <140 | 89.3 | 85.7 | 14.3 | 10.7 |

Fig. 4.1 shows the example outputs of skin like region segmentation of different test images with the three given algorithms. It can be observed that all the algorithms were able to detect the majority of the skin area in the images.

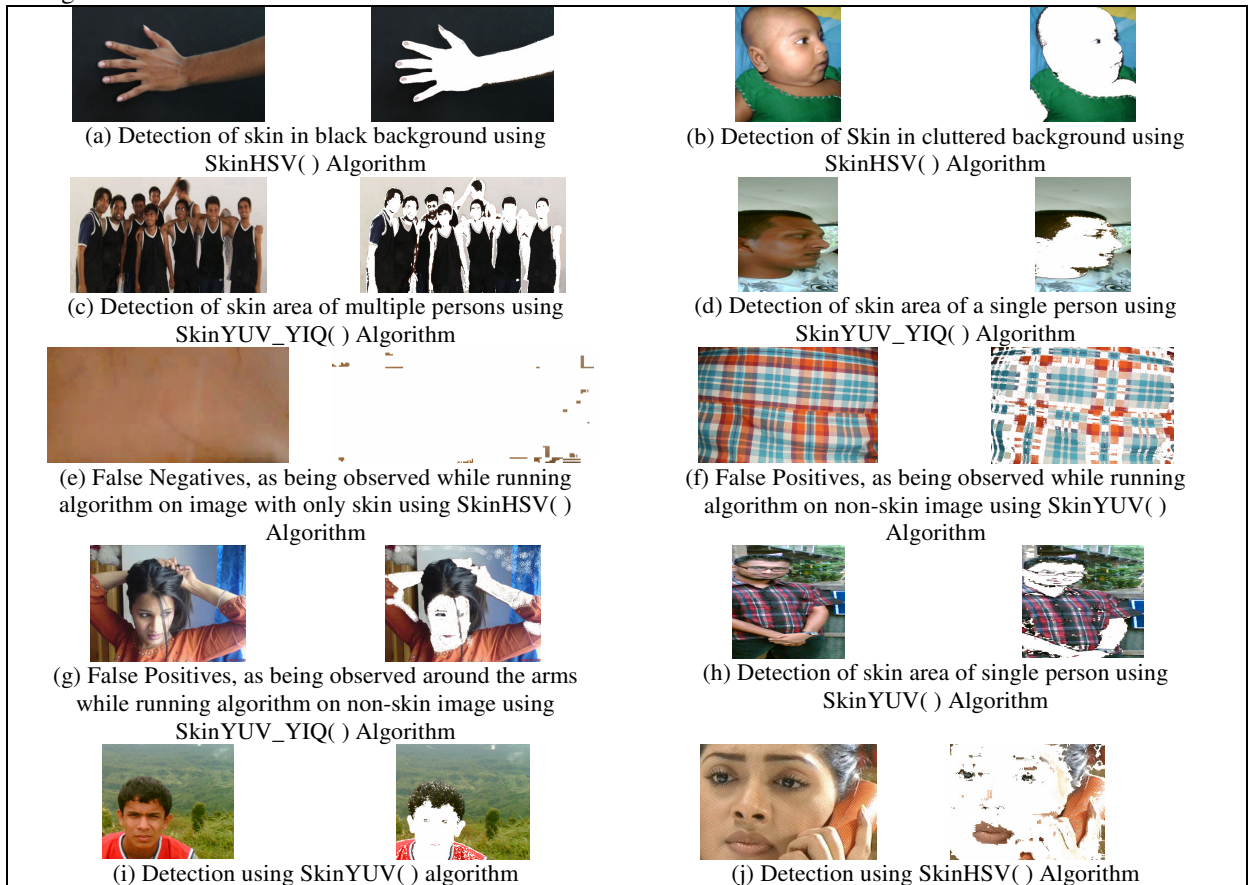

(a) Detection of skin in black background using SkinHSV( ) Algorithm

(b) Detection of Skin in cluttered background using SkinHSV( ) Algorithm

(c) Detection of skin area of multiple persons using SkinYUV_YIQ( ) Algorithm

(d) Detection of skin area of a single person using SkinYUV_YIQ( ) Algorithm

(e) False Negatives, as being observed while running algorithm on image with only skin using SkinHSV( ) Algorithm

(f) False Positives, as being observed while running algorithm on non-skin image using SkinYUV( ) Algorithm

(g) False Positives, as being observed around the arms while running algorithm on non-skin image using SkinYUV_YIQ( ) Algorithm

(h) Detection of skin area of single person using SkinYUV( ) Algorithm

(i) Detection using SkinYUV( ) algorithm

(j) Detection using SkinHSV( ) Algorithm

**Fig. 4.1:** Example of Skin Detection

As can be observed from the figure, there are some areas of non skin images detected as skin by the algorithms, these are False Positives. There are, however, some skin areas in some images that could not be rightly detected as skin by the given algorithms, these are False Negatives. The graphs in Fig. 4.2 - 4.5 present the performance of the three discussed algorithms on randomly chosen 20 images.

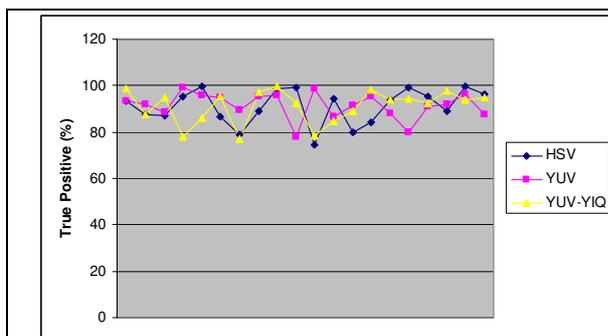

**Fig. 4.2:** True positive for three algorithms

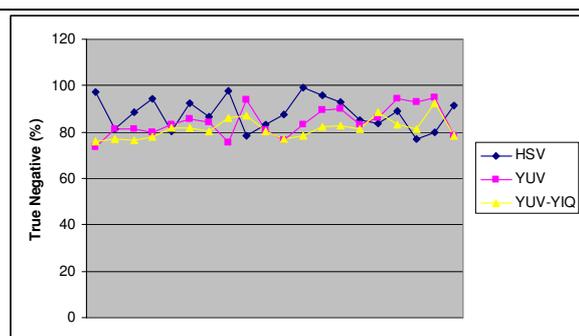

**Fig. 4.3:** True negative for three algorithms

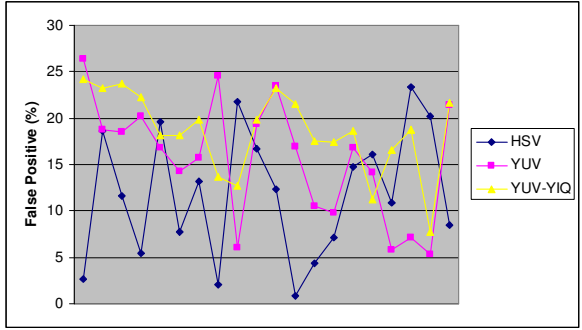 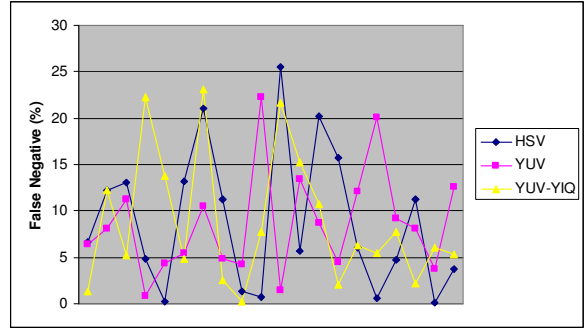

**Fig. 4.4:** False positive for three algorithms     **Fig. 4.5:** False negative for three algorithms

Table 4.4 Summarizes the performance of given algorithms tested on 80 images with optimized thresholds with respect to success parameters – true positive, true negative, false positive, false negative.

**Table 4.4:** Average Performance

| Algorithm | True Positive(%) | True Negative(%) | False Positive(%) | False Negative(%) |
|---|---|---|---|---|
| HSV | 91.1 | 88.1 | 11.9 | 8.9 |
| YUV | 91.4 | 84.4 | 15.6 | 8.6 |
| YUV & YIQ | 91.2 | 81.5 | 18.5 | 8.8 |

From the above table, considering the outcome on experimentation on three different skin detection algorithms, trained using 120 images and tested on 80 separate images, we see that HSV model based approach to Indian sub-continental skin recognition showed the best performance.

## 5. Conclusion

With the development of World Wide Web, there has been a dramatic advancement in the world of photography as a result of effective coding mechanisms and dazzling array of image models using certain color space. These models are extensively used to detect skin and to find out the pattern in terms of pixel based skin detection. The historical background of image processing is highly motivating and blessed with some tremendous research works. This paper is a reflection of the color space models that are used in this area.

This paper made a comparison between the three skin detection approaches based on three different color models. It presents a statistical analysis to evaluate the performance of the methods with respect to success measures like true positive, false negative, true negative and false positive. This study has tried to explore which of the three methods is best suited for detection of sub-continental skins. Among the various methods, the HSV color model based method with 91.1% true positive 88.1% true negative is proved to be best suited for detecting human skin of Indian sub-continental males and females. The testing also reflected certain flaws which were rectified by optimization of detecting parameters. In the next stage of research combinatorial approach of using multiple color models by boosting of given algorithms can be tested for better detection rates.

**Acknowledgements:** This work was supported in part by the University of Dhaka.